\documentclass[sn-chicago]{sn-jnl}

\usepackage{setspace}
\usepackage{moresize}

\usepackage{graphicx}%
\usepackage{amsmath,amssymb,amsfonts}%
\usepackage{amsthm}%
\usepackage{mathtools}
\usepackage{mathrsfs}%
\usepackage[title]{appendix}%
\usepackage{xcolor}%
\usepackage{textcomp}%
\usepackage{manyfoot}%
\usepackage{booktabs}%
\usepackage{algorithm}%
\usepackage{algorithmicx}%
\usepackage{algpseudocode}%
\usepackage{listings}%

\usepackage{hyphenat}

\usepackage{cleveref}
\usepackage{moresize}
\usepackage{bm}
\usepackage{enumerate}
\usepackage{tabularx}
\usepackage{makecell}
\usepackage{multirow}
\usepackage{tablefootnote}
\usepackage{tabularray}
\UseTblrLibrary{booktabs}
\usepackage{xfrac} 
\usepackage{xifthen}
\usepackage{txfonts}

\usepackage{caption}
\usepackage{subcaption}

\bibpunct
{(} 
{)} 
{,} 
{a} 
{} 
{;} 

\newtheorem{definition}{Definition}%

\def\argmax{\mathop{\mathrm{arg}\,\mathrm{max}}}                                          
\def\argmin{\mathop{\mathrm{arg}\,\mathrm{min}}}                                          
\newcommand{\rv}[1]{#1}
\newcommand{\rvs}[1]{\boldsymbol{#1}}
\newcommand{\set}[1]{\mathcal{#1}} 
 \newcommand{\prob}{\mathbb{P}}
\def\cond{\,|\,}

\newcommand{\nl}{\newline\hspace*{5mm}}
\definecolor{greenish}{RGB}{0, 153, 76}
\definecolor{grayish}{RGB}{140, 140, 140}

\begin{document}

\title[Scientific Inference With Interpretable Machine Learning]{\large{Scientific Inference With Interpretable Machine Learning}\linebreak \small{Analyzing Models to Learn About Real-World Phenomena}}

\author*[1]{\fnm{Timo} \sur{Freiesleben}}\email{timo.freiesleben@uni-tuebingen.de}

\author[1]{\fnm{Gunnar} \sur{König}}

\author[2]{\fnm{Christoph} \sur{Molnar}}

\author[1]{\fnm{Alvaro} \sur{Tejero-Cantero}}

\affil*[1]{\orgdiv{Cluster of Excellence Machine Learning for Science}, \orgname{University of Tübingen}, \orgaddress{\street{Maria-von-Linden-Straße 6}, \city{Tübingen}, \postcode{72076}, \country{Germany}}}

\affil[2]{\orgdiv{Independent Researcher}, \city{Munich}, \country{Germany}}



\abstract{To learn about real world phenomena, scientists have traditionally used models with clearly interpretable elements. However, modern machine learning (ML) models, while powerful predictors, lack this direct elementwise interpretability (e.g. neural network weights). Interpretable machine learning (IML) offers a solution by analyzing models holistically to derive interpretations. Yet, current IML research is focused on auditing ML models rather than leveraging them for scientific inference. Our work bridges this gap, presenting a framework for designing IML methods---termed 'property descriptors'---that illuminate not just the model, but also the phenomenon it represents. We demonstrate that property descriptors, grounded in statistical learning theory, can effectively reveal relevant properties of the joint probability distribution of the observational data. We identify existing IML methods suited for scientific inference and provide a guide for developing new descriptors with quantified epistemic uncertainty. Our framework empowers scientists to harness ML models for inference, and provides directions for future IML research to support scientific understanding.}


\maketitle
\section{Introduction}
\label{sec:intro}
Scientists increasingly use machine learning (ML) in their daily work. This development is not limited to natural sciences like material science \citep{schmidt2019recent} or the geosciences  \citep{reichstein2019deep}, but extends to social sciences such as educational science \citep{luan2021review} and archaeology \citep{bickler2021machine}.
\nl
When building \emph{predictive} models for problems with complex data structures, ML outcompetes classical statistical models in both performance and convenience. Impressive recent examples of successful prediction models in science include the automated particle tracking at CERN \citep{farrell2018novel}, or DeepMind's AlphaFold, which has made substantial progress in predicting protein structures from amino acid sequences \citep{senior2020improved}. In such examples, some see a paradigm shift towards theory-free science that ``lets the data speak'' \citep{anderson2008end,mayer2013big,kitchin2014big,Spinney2022}. However, purely prediction driven research has its limits: In a survey with more than 1,600 scientists, almost 70\% expressed the fear that the use of ML in science could lead to a reliance on pattern recognition without understanding \citep{van2023ai}. This is in line with the philosophy of science literature, which does recognize the importance of predictions \citep{douglas2009reintroducing,luk2017theory}, but also emphasizes other goals such as explaining and understanding phenomena \citep{salmon1979ask,shmueli2010explain,longino2018fate,toulmin1961foresight}.
\nl
The reason why understanding phenomena with ML is difficult is that, unlike traditional scientific models, ML models do not provide a cognitively accessible representation of the underlying causal mechanism \citep{hooker2017machine,boge2022two,molnar2024MLscience}. 
The link between the ML model and phenomenon is unclear, leading to the so-called \emph{opacity problem} \citep{sullivan2022understanding}.
Interpretable machine learning (IML, also called XAI, for eXplainable artificial intelligence) aims to tackle the opacity problem by analyzing individual model elements or inspecting specific model properties \citep{molnar2020interpretable}. However, it often remains unclear how IML can help to address problems in science \citep{roscher2020explainable} as IML methods are often designed with other purposes and stakeholders in mind, such as guiding engineers in model construction or offering decision support for end users \citep{zednik2021solving}.
\nl
In spite of these challenges, scientists increasingly use IML for inferring which features are most predictive of crop yield \citep{shahhosseini2020forecasting,zhang2019california}, personality traits \citep{Stachl17680}, or seasonal precipitation \citep{gibson2021training}, among others.
Although researchers are aware that their IML analyses remain just model descriptions, it is often implied that the explanations, associations, or effects they uncover extend to the corresponding real-world properties. 
Unfortunately, drawing inferences with current IML raises epistemological issues \citep{molnar2022pitfalls}: it is often unclear what target quantity IML methods estimate \citep{lundberg2021your}, is it properties of the model or of the phenomenon \citep{chen2020true,hooker2021unrestricted}? Moreover, theories to quantify the uncertainty of interpretations are underdeveloped \citep{molnar2020history,watson2022conceptual}.

\paragraph{Contributions}
In this paper, we present an account of scientific inference with IML inspired by ideas from philosophy of science and statistical inference. We focus on supervised learning on identically and independently distributed (i.i.d.) data relating predictors $\rvs{X}$ and a target variable $\rv{Y}$. Our key contributions are: 
\begin{itemize}
    \item We argue that ML cannot profit from the traditional approach to scientific inference via model elements because individual ML model parameters do not represent meaningful phenomenon properties.  We observe that current IML methods generally do not offer an alternative route: while some do interpret the model as a whole beyond its elements, they are designed to support model audit and not scientific inference.
    \item We identify the criteria that IML methods need to fulfill so that they provide access to the properties of the conditional probability distribution $\prob(\rv{Y}\cond\rvs{X})$. We provide a constructive approach to derive new IML methods for scientific inference (we call them \emph{property descriptors}), and identify for which estimands existing IML methods can be appropriated as property descriptors. We discuss how property descriptions can be estimated with finite data and quantify the resulting epistemic uncertainty.
\end{itemize}

\paragraph{Roadmap}
This paper is addressed to an interdisciplinary audience, where various communities may find some sections of special interest:
\begin{itemize}
    \item Philosophers of science may start with our discussion in \Cref{sec:traditionalModeling} on traditional scientific modeling and why its notion of representation is not longer applicable to ML models (see~\Cref{sec:RepresentationalityML} and \Cref{sec:concepts}). Additionally, \Cref{sec:causality} offers insights on what causal understanding can be derived from ML model interpretations. 
    \item  Researchers in IML may want to skip ahead to \Cref{sec:steps}, where we offer a concise guide to selecting or developing IML methods apt for scientific inference. In preparation, sections \Cref{sec:IML} and \Cref{sec:HR} motivate our approach and \Cref{subsec:WhatMLmodelsRepresent} reviews some necessary mathematical background. We suggest future avenues for IML research in~\Cref{sec:discussion}.
    \item Finally, scientists will find in~\Cref{tab:CommonlyRelevantDes} a list of published IML methods that address a variety of practical inference questions, with~\Cref{sec:discussion} discussing limitations to consider in applications.
\end{itemize}

\paragraph{Terminology}
For the purposes of our discussion below, a \emph{phenomenon} is a real-world process whose \emph{aspects of interest} can be described by random variables; these aspects possess distinct \emph{properties}. Observations of the phenomenon are assimilated to draws from the unknown joint distribution of the random variables to form a dataset, or just \emph{data}. \emph{Scientific inference} describes the process of rationally inducing knowledge about a phenomenon from such observational data (via ML, or other types of models), providing a  basis for \emph{scientific explanations}. In this work, when we talk about ML, we focus exclusively on the supervised learning setting. A supervised \emph{ML model} is a mathematical function with some free parameters that a learning algorithm optimizes (``trains'') based on existing labeled data in order to accurately predict future or withheld observations, i.e. to generalize beyond the initial data. While these predictions are often called `inferences' in the deep learning literature, we here employ \emph{inference} in its original sense in statistics: investigating unobserved variables and parameters to draw conclusions from data. In this sense, inference goes beyond prediction; it concerns the data structure and its uncertainties. 
When we talk about IML methods in this paper, we include all approaches that analyze trained ML models and their behavior. 
In contrast to \cite{rudin2019stop}, we do not limit the scope of IML to inherently interpretable models. 
These brief conceptual remarks are meant to reduce ambiguity in our usage, we lay no claim as to their universality.

\section{Related work}
\label{sec:Background}
Whether and how ML models, and specifically IML, can help obtain knowledge about the world is a debated topic in philosophy of science, statistics, causal inference, and within the IML community.

\paragraph{Philosophy of science} According to  \cite{bailer2002modeling} and \cite{bokulich2011scientific} ML models are only suitable for prediction, since their parameters are merely instrumental and lack independent meaning. Conversely, \cite{sullivan2022understanding} contends that nothing prevents us from gaining real-world knowledge with ML models as long as the \emph{link uncertainty}---the connection between the phenomenon and the model---can be assessed. While \cite{cichy2019deep} and \cite{zednik2022scientific} claim that IML can help in learning about the real world, they remain vague about how model and phenomenon are connected. We agree with \cite{watson2022conceptual} that only IML methods that evaluate the model on realistic data can allow insight into the phenomenon, but clarify that without further assumptions such methods only reveal associations learned by the ML model, not causal relationships in the world \citep{raz2022understanding,kawamleh2021can}. Our work makes precise that ML models can be described as epistemic representations of a certain phenomenon that allow us to perform valid inferences \citep{contessa2007scientific} via interpretations.

\paragraph{Statistical modeling and machine learning}
\cite{breiman2001statistical} distinguishes between algorithmic (ML) and data (statistical) modeling. He illustrates, using a medical example, that interpreting high-performance ML models post-hoc can yield more accurate insights about the underlying phenomena compared to standard, inherently interpretable data models.
Our paper provides an epistemic foundation for such post-hoc inferences. \cite{shmueli2010explain} distinguishes statistics and ML by their goals---prediction (ML) and explanation (statistics). Like \cite{hooker2021bridging}, we argue against such a clear distinction and offer steps to integrate the two fields.
To do so, we initially clarify what properties of a phenomenon can be addressed by IML in principle. Starting from theoretical estimands, we develop a guide for constructing IML methods for finite data, following the scheme in \cite{lundberg2021your}. Finally, based on the approach by \cite{Molnar2023relating}, we show how to obtain uncertainty estimates for IML-based inferences.

\paragraph{Causal inference using machine learning}
ML is now widely used as a tool for estimating causal effects \citep{kunzel2019metalearners,knaus2022double,dandl2023causality}. \emph{Double machine learning}, for example, provides an ML-based unbiased and data-efficient estimator of the (conditional) average causal effect, assuming that all confounding variables are observed \citep{chernozhukov2018double}.\footnote{The double ML methodology has also been applied for other inferential problems, see \cite{fink2023double}.} While these works focus on the construction of practical estimators for causal effects, we focus on the question \emph{what} inferences can be drawn from interpreting individual ML models and \emph{how} to design property descriptors for such inferences.
\nl
The Targeted Learning framework by \citep{van2011targeted,van2018targeted} is closely related to our proposal and also to double ML \citep{diaz2020machine,hines2022demystifying}. Like us, they discourage the use of interpretable but misspecified parametric models for scientific inference. Instead, they suggest to directly estimate relevant properties (they call them \emph{parameters}) of the joint data distribution that are motivated by causal questions. They estimate relevant aspects of the joint distribution from which the parameters can be derived using the so-called \emph{super learner} (a weighted ensemble of ML models), and debias their estimator with the targeted maximum likelihood update \citep{hines2022demystifying,van2006targeted}. Compared to targeted learning, we ask more specifically what inferences we can draw from interpreting individual ML models and how to match such interpretations with parameters of traditional scientific models. Also, we primarily provide a theoretical framework for IML tools and how they must be designed to gain insights into the data, whereas the work of \cite{van2011targeted} is more practical, providing unbiased estimators and implementations for a variety of inference tasks \citep{van2006targeted}. We further discuss the relationship between property descriptors and targeted learning in \Cref{sec:discussion}.

\paragraph{Interpretable machine learning}
IML as a field has been widely criticized for being ill-defined, mixing different goals (e.g. transparency and causality), conflating several notions (e.g. simulatability and decomposability), and lacking proper standards for evaluation \citep{doshi2017towards,lipton2018mythos}. Some even argued against the central IML leitmotif of analyzing trained ML models post hoc in order to explain them \citep{rudin2019stop}. In this paper, we show that if we focus solely on interpretations for scientific inference, we can clearly define what estimand post hoc IML methods estimate and how well they do so.
\nl
Using IML for scientific inference is not a completely new idea. It has been proposed for IML methods like Shapley values \citep{chen2020true}, global feature importance \citep{fisher2019all} and partial dependence plots \citep{Molnar2023relating}. Our framework generalizes these method-specific proposals to arbitrary IML methods and provides guidance on how to construct IML methods that enable inference.

\section{The traditional approach to scientific inference requires model elements that meaningfully represent}
\label{sec:traditionalModeling}
\begin{figure}[h]
\centering
  \includegraphics[width=0.85\linewidth]{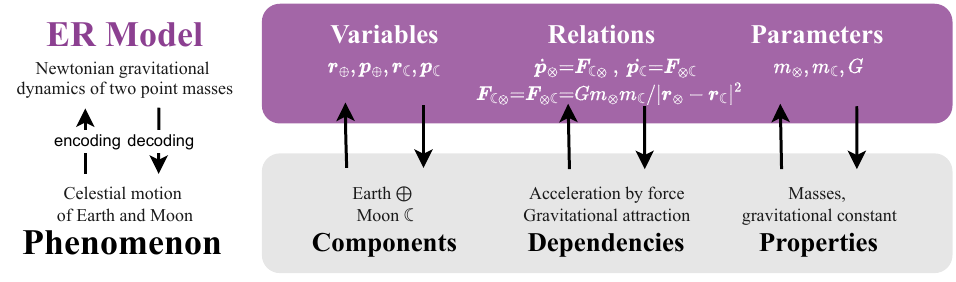}
  \caption{\textbf{Model and phenomenon sustain an encoding-decoding relationship}.  The main elements of a traditional, ER model, are shown in encoding-decoding correspondence to the phenomenon elements they represent  \citep{stachowiak1973allgemeine,contessa2007scientific}. Phenomenon and model elements are illustrated with a simple example of two bodies in gravitational interaction and its classical, Newtonian mechanistic description. This physical example was chosen to illustrate the ER paradigm, we make no claim that our property descriptors presented later will achieve similar representational power.}

  \label{fig:PhenModel}
\end{figure}

Models can be found everywhere in science, but what are they really? Like Bailer-Jones, we see a scientific model as an ``interpretative description of a phenomenon that facilitates perceptual as well as intellectual access to that phenomenon'' \cite[p61]{bailer2003scientific}. The way models provide access to phenomena is usually specified by some sort of representation \citep{sep-scientific-representation,sep-models-science}. Indeed, models represent only some aspects of a phenomenon, not all of it---a good model is true to the aspects that are relevant to the model user \citep{bailer2003scientific,ritchey2012outline}.
\nl
In scientific modeling, we noted a paradigm that many models implicitly follow---we call it the paradigm of \emph{elementwise representationality}.
\begin{definition}
A model is \emph{elementwise representational} (ER) if all model elements (variables, relations, and parameters) represent an element in the phenomenon (components, dependencies, properties). 
\label{def:ER}
\end{definition}
\Cref{fig:PhenModel} illustrates how models relate to the phenomenon when they are ER (in the example, two-body Newtownian dynamics is used to model the Earth-Moon system) : variables describe phenomenon components; mathematical relations between variables describe structural, causal or associational dependencies between components; parameters specify the mathematical relations and describe properties, like the strength, of the component dependencies. By distinguishing components, dependencies and properties, we closely follow inferentialist accounts of representation by \cite{contessa2007scientific} and other philosophers like \cite{Achinstein1968-ACHCOS,stachowiak1973allgemeine,hughes1997models,bailer2003scientific,levy2012models} and \cite{ducheyne2012scientific}. The upward arrows in \Cref{fig:PhenModel} describe \emph{encoding} into representations i.e. the translation of a phenomenon observation to a model configuration;  The downward arrows describe \emph{decoding} i.e. the translation of knowledge about the model into knowledge about the phenomenon.

ER does not mean that each model element represents independently of the rest of the model. Instead, ER is model-relative. When we specify the rest of the model, ER implies that each model element has a fixed meaning in terms of the phenomenon. We also want to emphasize that ER is an extreme case. There are indeed models in science where not every model element represents but some parts of the overall model do. A typical non-representational element used in scientific models is a noise term. Instead of representing a specific component
or a property of the phenomenon, the noise can be a placeholder for unexplained aspects of the
phenomenon \citep{edmonds2006nature}.
\nl
ER is obtained through model construction; ER models are usually ``hand-crafted'' based on background knowledge and an underlying scientific theory. Variables are selected carefully and sparsely during model construction, and relations are constrained to a relation class with few free parameters. When ER models need to account for additional phenomenon aspects, they are gradually extended so that large parts of the ``old'' model are preserved in the more expressive ``new'' model \citep{schwarz2009developing}. ER even eases this model extension process because model interventions are intelligible on the level of model elements. Usually, ER is explicitly enforced in modeling: if there is a phenomenon element devoid of meaning, researchers either try to interpret it or exclude it from the model.
\nl
ER is so remarkable because it gives models capabilities that go beyond prediction. ER has a crucial role in translating model knowledge into phenomenon knowledge \citep[\emph{surrogative reasoning}][]{swoyer1991structural,hughes1997models,contessa2007scientific}. Scientists can analyze model elements and draw immediate conclusions about the corresponding phenomenon element \citep{sep-scientific-representation}. However, only those properties of the phenomenon that have a model counterpart can be analyzed with this approach. Fortunately, as described above, ER models can be extended to account for further relevant phenomenon elements identified by the scientist.

\paragraph{Example associational ER model: simple linear regression}
The mechanistic causal model from \Cref{fig:PhenModel} is ideal for illustrating what constitutes an ER model. However, ML models are associational in nature. To better isolate the differences in the ways ER and ML models represent, we now focus on an associational ER model, in this case a simple linear model.
\nl
 Suppose we want to study which factors influence students' skills in math, specifically focusing on language mastery \citep{peng2020examining}. We adopt grades as quantitative proxies for the respective skills, and find in \cite{cortez2008using} a suitable dataset reporting Portuguese and math grades on a 0-20 scale, along with 30 other variables such as student age, parents' education, etc (see Appendix~\ref{app:dataset} for details). Here, the students' characteristics like their math and Portuguese skills are the phenomenon components, genetic and environmental associations are the dependencies, and the strength or direction of these associations are instances of relevant properties. 
\nl
We start with a linear regression with the Portuguese grade as the only predictor variable, denoted $\rv{X}_p$, and the math grade as the target, $\rv{Y}=\beta_0+\beta_1 \rv{X}_p+\epsilon$, with $\beta_0,\beta_1\in\mathbb{R}$. This linear relation is a reasonable, if crude, assumption for a first analysis in absence of prior insight.\footnote{Note that our toy model is for illustration only, and not meant to reflect social science methodology.} Our model is ER except for the noise term, $\epsilon\sim \mathcal{N}(0,\sigma^2)$, which accounts for all variability in $\rv{Y}$ not due to $\rv{X}_p$ and thus is, by design, not ER.
We train the model by finding the $\hat{\beta}_1,\hat{\beta}_0$ that minimize the mean-squared-error (MSE) of predictions on the training set:
\begin{equation}
    \hat{m}_{\rm LIN}(x_p)= \hat{\beta}_0 + \hat{\beta}_1 x_p. \label{eq:linearModel}
\end{equation} 
The fitted regression coefficient $\hat{\beta}_1=0.77$ has a 95\% confidence interval (CI) of $[0.63<\beta_1<0.91]$,\footnote{This means that the proportion of CIs (each calculated from a theoretical, newly-sampled dataset) that contain the true parameter value $\beta$ tends in the long-run to 95\% \citep{heumann2016introduction}.} and represents the association as the expected increase in math grade for a unit increase in Portuguese grade. The bias $\hat{\beta}_0$ is also straightforward to interpret.\footnote{The expected math grade for a zero in Portuguese is $\hat{\beta}_0=0.80$. However, regressing on $x_p-\bar{x}_p$ instead gives the more useful expected grade of an average Portuguese student (with $\bar{x}_p=12.55$), $\hat{\beta}^{\rm avg}_0=10.46$, with a 95\% CI of $10.05<\beta^{\rm avg}_0<10.88$.} Thus, from $\hat{\beta}_1$ and its CI, we might conclude with some confidence that language and math skills are positively and strongly associated. This conclusion is contingent on the model being ER, but crucially, also on it capturing well the phenomenon. Although by optimizing the MSE we targeted an appropriate estimand, namely the conditional expectation $\mathbb{E}_{\rv{Y}\cond \rv{X}_p}[\rv{Y}\cond \rv{X}_p]$, we estimated it with a rather crude model. Clearly, if the model is not highly predictive, it is ill-suited for reliable scientific inference. To improve performance, we can make the model more expressive by introducing additional variables, relations, or interaction parameters. As long as we preserve ER, we can draw scientific inferences directly from model elements. These inferences are only as valid as the modeling assumptions (e.g. target normality, homoscedasticity, or linearity).

\section{The elements of ML models do not meaningfully represent}
\label{sec:RepresentationalityML}
ER models suit our image of science as an endeavor aimed at understanding. ER enables us to reason about the phenomenon, and in causal models it even allows us to reason about effects of model or even real-world interventions.  However, when constructing ER models, we require background knowledge about which components are relevant, and we usually need to severely restrict the class of relations that can be considered in modeling a given phenomenon. These difficulties might lead scientists to either limit their investigations to simple phenomena or to settle for overly simplistic models for complex phenomena and, as \cite{breiman2001statistical} and \cite{van2011targeted} cautioned, possibly draw wrong conclusions.
\nl
ML models, on the other hand, excel for complex problems with an unbounded number of components that entertain ambiguous and entangled relationships, i.e. ML models are highly expressive \citep{guehring_raslan_kutyniok_2022}. Indeed, given enough data, effective prediction with ML requires less subject-domain background knowledge than traditional modeling approaches \citep{bailer2002modeling}. 
While the definition of a prediction task and the encoding of features are still largely based on domain knowledge, the choice of model class, hyperparameters and learning algorithms is often guided by the data types and aims to promote efficient learning for the respective modality, e.g. by selecting architectures such as CNNs for images, LSTMs for sequences or GNNs for relational data.
\nl
The gain in generality and convenience with ML comes at a price---ML models are generally far from being ER. As also argued in \cite{bailer2002modeling,bokulich2011scientific} and \cite{boge2022two}, ML models (e.g. paradigmatically artificial neural networks) contain model elements such as weights that have no obvious phenomenon counterpart.

\paragraph{Example ML associational model: artificial neural network (ANN)} 
Suppose we want to improve on the predictive performance of our simple linear model (Eq.~\ref{eq:linearModel}), and opt instead for a dense three-layer neural network that uses all available features to predict math grades. To train it, we minimize the MSE on training data, but now use gradient descent with an adaptive learning rate for lack of an analytical solution. The trained model can be described as a function parameterized by $31\times31$ weight matrices $\hat{W}_1$, $\hat{W}_2$, $\hat{W}_3$ and $31\times 1$ bias vectors $\hat{\bm{b}}_1,\hat{\bm{b}}_2,\hat{\bm{b}}_3$ using component-wise ReLU activations $\operatorname{ReLU}(\bm x_j) := \max(\bm{x}_j,0)$:
\begin{equation}
    \hat{m}_{\mathrm{ANN}}(\bm{x})=\hat{W}_3\,\operatorname{ReLU}\Bigl(\hat{W}_2\,\operatorname{ReLU} \bigl(\hat{W}_1 \bm{x}+\hat{\bm{b}}_1\bigr)+\hat{\bm{b}}_2\Bigr)+\hat{\bm{b}}_3. \label{eq:ANNModel}
\end{equation}
 While this ANN achieves a test-set MSE of just $8.9$ compared to $16.0$ for the simple linear model,\footnote{Using a multiple linear model for fairer comparison results in a MSE of $12.4$, reflecting a limited capacity to capture complex relationships.} it is now highly unclear what aspects of our phenomenon the ANN parameters correspond to. Its weights and biases are very hard or even impossible to interpret: any individual weight might have a positive, neutral, or negative effect on the target, dependent on all other model elements. Similarly, the design of the architecture aims to maximize predictive performance, rather than to reflect any actual or even hypothesized dependencies between components of the phenomenon.

\section{But do ML model elements really not represent?}
\label{sec:concepts}
\begin{figure}[h]
\centering

   \includegraphics[width=0.75\linewidth]{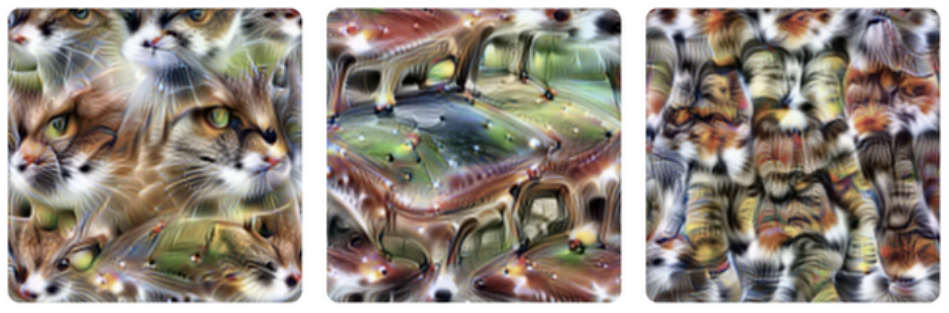}

    \caption{\textbf{ML models are generally not ER.} Three input images synthesized to maximally activate a given unit in a neural network \citep{olah2020zoom} illustrate how ``concepts'' as different as cat faces, fronts of cars, or cat legs all elicit strong responses, suggesting neural network elements generally do not represent unique concepts \citep{mu2020compositional,nguyen2016multifaceted}.} 
  \label{fig:neuronAct}
\end{figure}

One may believe that we went too fast here and argue that individual elements of ML models do have a natural phenomenon counterpart, one that only surfaces after extensive scrutiny. The underlying intuition is that human representations are near-optimal to perform prediction tasks and will be eventually rediscovered by ML algorithms. We find this unlikely: ER is not enforced in most state-of-the-art models \citep{leavitt2020selectivity} and, even worse, some widely-employed ANN training techniques, such as \emph{dropout}, purposefully discourage ER in order to gain robustness \citep{srivastava2014dropout}. High-capacity ML models like ANNs are indeed designed for \emph{distributed representation} \citep{sep-connectionism,mcclelland1987parallel}.
\nl
Still, it has been claimed that model elements represent high-level constructs constituted from low-level phenomenon components that are often called \emph{concepts} \citep{buckner2018empiricism,olah2020zoom,bau2017network,raz2023methods}. The idea is that similar to the hierarchical schema we use to understand nature, with lower level components such as atoms combining to form higher level entities such as molecules, cells, and organisms, representations in deep nets evolve through layers from pixels to shapes to objects. If this were the case, model elements or aggregates of such elements could be reconnected to the respective phenomenon entities; ER would be restored by the representations of coarse-grained phenomenon components.
\nl
While empirical research on neural networks finds that some model elements are weakly associated with human concepts \citep{mu2020compositional,voss2021visualizing,kim2018interpretability,olah2017feature,bau2017network,bills2023language}, often these elements are neither the only associated elements nor exclusively associated with one concept as shown in \Cref{fig:neuronAct}  \citep{donnelly2019interpretability,bau2017network,olah2020zoom}. Moreover, intervening on these model elements generally does not have the expected effect on the prediction---the elements do not share the causal role of the ``represented'' concepts, even in prediction \citep{GALE202060,donnelly2019interpretability,zhou2018revisiting}. It is therefore questionable in what sense, or even whether, individual elements of ML models represent \citep{freiesleben2023artificial}. Moreover, this line of research that tries to identify concepts in model elements predominantly focuses on images, where nested concepts are arguably easy to identify for humans. 

 In sum, current ML algorithms do not enforce ER---hence, trained ML models rely on distributed representations and cannot be reduced to logical concept machines. An associative connection between a model element and a phenomenon concept should not be confused with their equivalence. While research on the representational correlates of model elements may indeed seem fascinating, analyzing single model elements appears to be a hopeless enterprise, at least if the goal is to support scientific inference.

\section{IML analyzes the model as a whole, but does it allow for scientific inference?}
\label{sec:IML}
Let us accept that ML models are indeed not ER. Can we still exploit their predictive power for scientific inference? We think this is indeed possible. Our approach is to regard the model as representational of phenomenon aspects \emph{only as a whole}---we call this \emph{holistic representationality} (HR). The idea behind HR is not new; it underlies, for example, modern causal inference \citep{van2011targeted}. HR has its roots in what \cite{heckman2000causal} calls \textit{Marschak’s Maxim} -- it is possible to directly describe an aspect of the data distribution without first identifying its individual components through a parametric model. ML models represent one paradigmatic case of HR models \citep{starmans2011models}.
\nl
The current research program in IML takes an HR perspective on ML models: Many IML methods analyze the entire trained ML model post-hoc just as an input-output mapping \citep{scholbeck2019sampling}, sometimes leveraging additional useful properties such as differentiability \citep{alqaraawi2020evaluating}.
\nl
Initial definitions of, for example, global feature effects \citep{friedman1991multivariate}, feature importance \citep{breiman2001random}, local feature contribution \citep{vstrumbelj2014explaining} or model behavior \citep{ribeiro2016should} have been presented. However, in recent years, many researchers have pointed out that these methods lead to counterintuitive results, and offered alternative definitions \citep{apley2020visualizing,strobl2008conditional,molnar2023model,goldstein2015peeking,konig2021relative,janzing2020feature,slack2020fooling,alqaraawi2020evaluating}. 
\nl
We believe that these controversies stem from a lack of clarity about the goal of model interpretation. Are we interested in model properties to learn about the model (model audit) or do we want to use them as a gateway to learn about the underlying phenomenon (scientific inference)? These two goals must not be conflated.
\nl
The auditor examines model properties e.g. for debugging, to check if the model satisfies legal or ethical norms, or to improve understanding of the model by intervening on it \citep{raji2020closing}. Another function that an audit can have is to assess the alignment between model properties and our (causal) background knowledge, such as certain monotonicity constraints \citep{tan2017auditing}, which is particularly important when ML is used in high-stakes decision making. In all those use-cases, auditors even take interest in model properties that have no corresponding phenomenon counterpart, such as single model elements or the behavior of the model when applied on unrealistic combinations of features. The scientist, on the other hand, wants to learn about model properties only in so far as they can be interpreted in terms of the phenomenon. 
\nl
This does not imply that model audit is irrelevant for scientists. In fact, a careful model audit that shows what the model does and how reliable it is appears as a \emph{prerequisite} for trustworthy scientific inference.

\section{How to think of scientific inference with HR models?}
\label{sec:HR}
\begin{figure}[h]
\centering
  \includegraphics[width=0.99\linewidth]{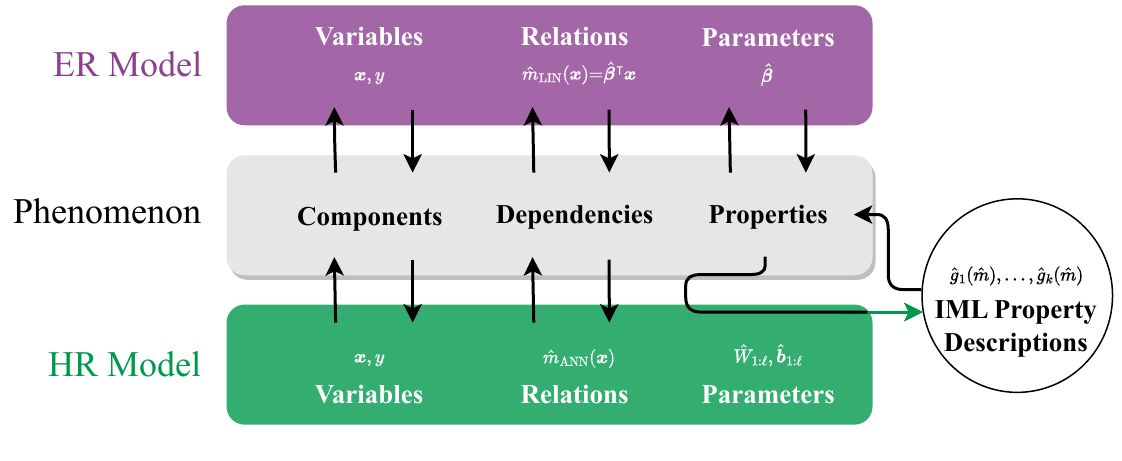}
  \caption{\textbf{Property descriptions distill phenomenon properties from HR models}. Instead of explicitly encoding phenomenon properties as parameters like for ER models, HR models (e.g. ML models) encode phenomenon properties in the whole model. We propose that these encoded properties can be read out with property descriptions external to the model. Property descriptors can take on the inferential role of parameters, for example of coefficients in linear models.
  }
  \label{fig:PhenModelNew}
\end{figure}

We just argued that ML models are generally not ER and therefore do not allow for scientific inference in the traditional way. HR offers a viable alternative, however, the IML research community currently conflates different goals of model interpretation. Our plan in the following sections is to show how a HR perspective enables scientific inference using IML methods. Particularly, we show which IML methods qualify as \emph{property descriptors}, i.e. quantify properties of the phenomenon, not just the model. \Cref{fig:PhenModelNew} describes our conceptual move: instead of matching phenomenon properties with model parameters as in ER models, we match them with external descriptions of the whole model.
\nl
In what follows, we will focus on scientific inference with trained ML models, as these constitute a paradigmatic and highly relevant category of HR models, even though our theory of property descriptors is generally applicable to any HR model as long as we know what it holistically represents.

In scientific inference, we start from a question concerning a phenomenon and some relevant data about that phenomenon. Even though in simple cases fitting properly constructed ER models can provide interpretable answers, for complex phenomena, a lack of model capacity often results in poor answers. In contrast, while recent ML models are opaque, they have the required representational capacity, and multiple IML methods already exist to probe them in various ways. What is missing, and what we are proposing here, is a way to map the initial question to a relevant IML method so that we can perform scientific inference even with models just aimed at predictive performance.
\nl
The key missing ingredient for scientific inference is to link the phenomenon and the model. We propose to draw this link using statistical learning theory, which characterizes what \emph{optimal} ML models can holistically represent. If the question can be in principle answered with an ideal model, we expect an approximate model to be able to answer it approximately. The problem is now reduced to qualifying what an approximate model is, and quantifying the error induced by the approximation. The same schema applies to using limited data as opposed to infinite data: we are fine with an approximate answer, so long as we can quantify the uncertainty.

\section{What aspects of phenomena do ML models holistically represent?}
\label{subsec:WhatMLmodelsRepresent}
Which aspects of a phenomenon ML models can represent under ideal circumstances depends on the data setting, the learning paradigm, and the loss function. We focus here on supervised learning from identically and independently distributed (i.i.d.) samples. In this widely useful setting, statistical learning theory provides us with optimal predictors \citep[p18-22]{hastie2009elements} as a rigorous tool to address model representation. 
\nl
Using the notation ${\rvs{X}\coloneqq(\rv{X}_1,\ldots,\rv{X}_n)}$ for the input variables and $\rv{Y}$ for the target variable with ranges $\set{X}$, and $\set{Y}$, we now formalize what characterizes the optimal prediction model and how to train such models from labeled data.

\paragraph{Optimal predictors}
An optimal predictor $m(\bm{x})$ predicts realizations of the target $\rv{Y}$ from realizations of the input $\rvs{X}$ with minimal expected prediction error i.e. $m=\underset{\hat{m}\in\set{M}}{\argmin}\,\mathrm{EPE}_{\rv{Y}|\rvs{X}}(\hat{m})$, with 
$$\mathrm{EPE}_{\rv{Y}|\rvs{X}}(\hat{m})\coloneqq \int_{\rv{Y}} L(\rv{Y},\hat{m}(\rvs{X}))\  \prob_{\rv{Y} | \rvs{X}}(y|\bm{x})\, {\rm d} y,$$
where $L$ is the loss function $L(\rv{Y},m(\rvs{X})): \set{X}\times \set{Y} \to \mathbb{R}^+$ and $\hat{m}$ a model in the class $\set{M}$ of mappings from $\set{X}$ to $\set{Y}$. \Cref{tab:OptimalPred} recapitulates the optimal predictors for some standard loss functions.

\begin{table}
\SetTblrStyle{caption-tag}{font=\bfseries}
\begin{talltblr}
[caption={\textbf{The optimal predictors for standard loss functions reflect aspects of $\prob(\rv{Y}\cond \rvs{X})$}.\\},
label={tab:OptimalPred},
headsep=6pt,
note{a}={Optimal predictors are, from top to bottom, the \emph{conditional} mean, median, mode, and pmf.}]
{ cells={valign=m}, rows={rowsep=4pt},
  width = 0.9\linewidth, vspan = even,
  row{1} = {font=\bfseries, b, c, gray!40, rowsep=5pt},
  colspec = {X[2.9,c] X[4.2,c] X[5,c] X[5,c]}, 
  cell{even}{1} = {r=2}{}, 
  cell{2-Z}{Y,Z} = {mode=math}, 
  vline{2,4} = {1-Z}{}, 
  hline{2} = {-}{},
  }
\toprule
Problem & Loss & $L(\rv{Y},\hat{m}(\rvs{X}))$  & Optimal predictor{\normalfont \TblrNote{a} } $m$ \\
\midrule
{Regression\\($\rv{Y}$ continuous)}
  & mean squared error&(\rv{Y}-\hat{m}(\rvs{X}))^2 & \mathbb{E}_{\rv{Y}\cond \rvs{X}}[\rv{Y}\cond \rvs{X}]\\
  & mean absolute error&\bigl|\rv{Y}-\hat{m}(\rvs{X})\bigr| & \operatorname{median}(\rv{Y}\cond \rvs{X})\\
\cmidrule[l]{1-Z} 
{Classification\\($\rv{Y}$ discrete)}
  &0-1 loss& 0\text{ if }\hat{m}(\rvs{X})=\rv{Y},\text{ else }1 & \argmax_{y\in\set{Y}}\prob(\rv{Y}{=}y\cond \rvs{X})\\
  & cross entropy&\sum_{r\in \set{Y}}\prob_{\rv{Y}}(r)\log \prob_{\hat{m}(\rvs{X})}(r) & \prob(\rv{Y}\cond \rvs{X}) \\
\bottomrule
\end{talltblr}
\end{table}

\paragraph{Supervised learning from finite data}
In supervised learning, we seek to approximate an optimal predictor $m$ by a model $\hat{m}$ based on a specific finite dataset ${\set{D}\coloneqq \{ (\bm{x}^{(1)},y^{(1)}),\dotsc,(\bm{x}^{(k)},y^{(k)})\}}$. This `training' is carried out by a learning algorithm $I:\Delta\rightarrow \set{M}$, which maps the class $\Delta$ of datasets of i.i.d. draws\footnote{The domain of $I$ is only completely specified when the parameters that define the learning procedure and the search space of the algorithm (called hyperparameters in the context of $\hat{m}$) are fixed. For our discussion, the reader may assume hyperparameters to have been fixed a priori by a human or an automated ML algorithm \citep{hutter2019automated}.}, $(\bm{x}^{(i)},y^{(i)})\sim (\rvs{X},\rv{Y}$), to a class of models $\set{M}$. Instead of the EPE itself, the learning algorithm minimizes the empirical risk on \textit{training data} and is then evaluated based on the empirical risk on \textit{test data} (i.e. data not contained in the training data), which constitutes a finite-data estimate of the EPE \citep{hastie2009elements}.

\section{Scientific inference with ML in four steps}
\label{sec:steps}
\begin{figure}[h]
\centering
   \includegraphics[width=0.95\linewidth]{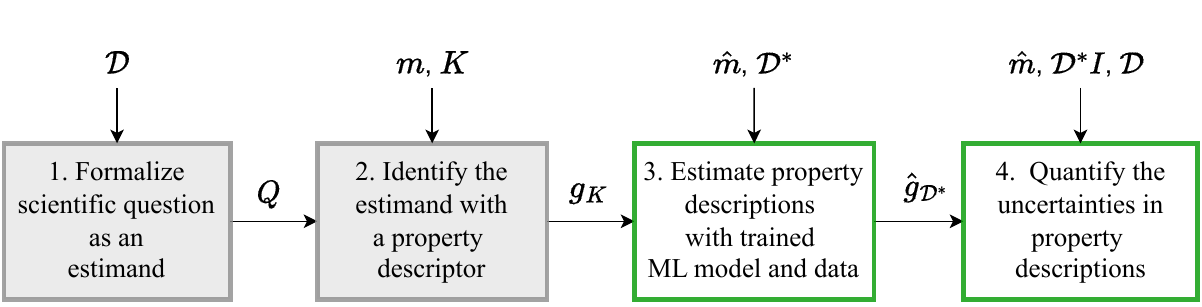}
  \caption{\textbf{An epistemic foundation for scientific inference with IML}. {\color{grayish}Steps 1 and 2} connect the property descriptors theoretically with an underlying estimand of scientific interest. {\color{greenish}Steps 3 and 4} show \emph{how} to practically draw inferences and quantify their uncertainty. See text for symbol definitions.}
  \label{fig:FourSteps}
\end{figure}

We have just outlined how (optimal) predictive models represent holistically aspects of the conditional distribution of the data, $\prob(\rv{Y}\cond \rvs{X})$, see \Cref{tab:OptimalPred}. In this section, we finally introduce \emph{property descriptors} as the tools that allow us to investigate these aspects by describing their relevant properties. Property descriptors exploit the ML model to study general associations in our data, providing insight beyond raw prediction. We provide a four-step plan to construct such descriptors that is illustrated in \Cref{fig:FourSteps}.

We assume the following scenario: interested in answering a scientific question about a specific phenomenon, a researcher seeks to exploit an informative labeled dataset about the phenomenon, $\mathcal{D}$ together with a highly predictive (ML) supervised model trained on it, $\hat{m}$. 
\nl
Our strategy encompasses the following key steps: 1. outlining the solution in an idealized context with an optimal predictor $m$ together with prior probabilistic knowledge $K$, rather than a real model and real data, 2. demonstrating the feasibility of approximating this ideal under certain assumptions, 3. estimating the solution from finite data and a trained model, and 4. quantifying the uncertainty inherent in these approximations.

\subsection*{Step 1: Formalize scientific question as an estimand}

The first step in our approach is to formalize the scientific question, $Q$, as a statistical \emph{estimand} \citep{lundberg2021your}. This estimand represents a probabilistic query on $\rvs{X}$ and $\rv{Y}$.

\paragraph{Example: the link between language and math skills}

An educator is interested in how math skills relate to language skills. She has access to a relevant labeled dataset, consisting of student grades in Portuguese and math, represented by the random variables $\rv{X}_p$ and $\rv{Y}$ respectively. The question of what is the expected grade in math for any given Portuguese grade can be formalized as an estimand, the conditional expectation, $Q:=\mathbb{E}_{\rv{Y}\cond \rv{X}_p}[\rv{Y}\cond \rv{X}_p]$.

\subsection*{Step 2: Identify the estimand with a property descriptor}

Having $Q$, we now ask if it can be derived from an \emph{optimal} model $m$ (e.g. one of those in \Cref{tab:OptimalPred}). Clearly, if $Q$ cannot be derived from $m$, using the actually available \emph{approximate} $\hat{m}$ will be inviable. Even $m$ alone will often not suffice. Since $m$ represents aspects of $\prob(\rv{Y}\cond \rvs{X})$, we also require probabilistic knowledge $K$ about $\prob(\rvs{X})$ and sometimes even of $\prob(\rv{Y})$ to derive relevant inferences from $m$. Note that $K$ is generally not available but must again be estimated from data. Following the reasoning with the ideal model, we consider whether the problem could be resolved assuming we have additional probabilistic knowledge $K$. 
\nl
We deem an estimand \emph{identifiable} with respect to available knowledge $K$ if it can be derived from the optimal predictor $m$ and $K$. To establish identifiability means to provide a constructive transformation of $m$ and $K$ into $Q$, ideally, the one with the most parsimonious requirements on $K$, since in the real world $K$ is obtained by collecting data or positing assumptions. We call this constructive transformation a \emph{property descriptor}, and demand that it be a continuous function $g_K$ w.r.t. metrics $d_{\set{M}}$ and $d_{\set{Q}}$:\footnote{The function $d_{\set{M}}$ is a metric on the function space $\set{M}$, $d_{\set{M}}(m_1,m_2)\coloneqq \int_{\rvs{X}}L(m_1(x),m_2(x))\,\prob_{\rvs{X}}(x)\ {\rm d}x$ for $m_1,m_2\in \set{M}$, while $d_{\set{Q}}$ describes a metric appropriate for the space $\set{Q}$.}
\begin{align*}
    g_K: \set{M} &\to \set{Q}\quad \text{ with } \quad g_K(m)=Q.
\end{align*}
The output space $\set{Q}$ remains deliberately unspecified; depending on the particular scientific question, we might want $\set{Q}$ to be a set of real numbers, vectors, functions, probability distributions, etc.

\paragraph{Example: property descriptor for a multivariable model}
Consider a multivariable model trained to minimize the MSE loss, such as our ANN (Eq.~\ref{eq:ANNModel}), which predicts math grades from \emph{all} available features. The model approximates the conditional expectation $m(\rvs{X})=\mathbb{E}_{\rv{Y}\cond \rvs{X}}[\rv{Y}\cond \rvs{X}]$.\footnote{Note the same discussion applies to a multiple linear regression, but, in contrast to the neural network, the coefficients of the linear model remain ER: the multiple regression coefficient $\hat{\beta}_j$ summarizes the additional contribution of $\rv{X}_j$ to change $\rv{Y}$ after $\rv{X}_j$ has been linearly adjusted for change in $\rvs{X}_{-i}$ \citep[in the data at hand;][sec. 3.2.3]{hoaglin2016regressions,hastie2009elements} to account for correlations among predictors.}
This is an optimal predictor, but in contrast to our desired estimand $Q$, it uses the Portuguese grade $\rv{X}_p$ as well as \emph{other features}, denoted $\rvs{X}_{-p}$. We can obtain $Q$ from $m$ by marginalizing over $\rvs{X}_{-p}$, i.e. our $Q$ is identifiable if we have the necessary $\prob(\rvs{X}_{-p}\cond \rv{X}_p):=K$ to compute the expectation\footnote{In the second line we use the \emph{tower rule} or \emph{rule of total expectation} whereby, for arbitrary random variables $\rvs{X},\rvs{Y},\rvs{Z}$, it holds $\mathbb{E}_{\rvs{Y}\cond\rvs{X}}[\rvs{Y}\cond\rvs{X}]=\mathbb{E}_{\rvs{Z}\cond\rvs{X}}\Bigl[\mathbb{E}_{\rvs{Y}\cond\rvs{X},\rvs{Z}}[\rv{Y}\cond\rvs{X},\rvs{Z}]\cond\rvs{X}\Bigr]$: it does not matter whether we directly take the expectation of $\rvs{Y}$ on $\rvs{X}$ or if we first take the expectation of $\rvs{Y}$ conditioned on a set of random variables $\rvs{X},\rvs{Z}$ that includes $\rvs{X}$ and then, ``integrate $\rvs{Z}$ out''.}
\begin{align*}
    Q & \coloneqq \mathbb{E}_{\rv{Y}\cond\rv{X}_p}[\rv{Y}\cond \rv{X}_p]\\ &=\mathbb{E}_{\rvs{X}_{-p}\cond\rv{X}_p}[\mathbb{E}_{\rv{Y}\cond\rvs{X}}[\rv{Y}\cond \rvs{X}]\cond \rv{X}_p] \qquad \text{(by the \emph{tower rule}})\\ &=\mathbb{E}_{\rvs{X}_{-p}\cond \rv{X}_p}[m(\rvs{X})\cond \rv{X}_p]\\
    &=g_K(m).
\end{align*}
Here $g_K$ is our property descriptor that takes $m$ and $K$ into $Q$. It is generally defined for an arbitrary model $\hat{m}\in\set{M}$ as
\begin{equation}
    g_K(\hat{m})(x_p)\coloneqq \mathbb{E}_{\rvs{X}_{-p}\cond\rv{X}_p}[\hat{m}(\rvs{X})\cond \rv{X}_p{=}x_p].
    \label{eq:cPDP}
\end{equation}
Note that $g_K(\hat{m})$ is continuous on $\set{M}$ and in fact corresponds to the well-known conditional partial dependence plot \citep[cPDP; also known as M-plot, ][]{molnar2020interpretable,apley2020visualizing}.

\subsection*{Step 3: Estimate property descriptions from a trained model and data}
In reality, we rarely have optimal predictors, they are theoretical constructs. Instead, we have trained ML models that approximate these theoretical constructs. We call the application of a property descriptor to our concrete ML model, $g_K(\hat{m})$, a \emph{property description}. Having assumed the continuity of property descriptors above guarantees that we obtain an approximately correct estimate when our ML model is close to the optimal model.
\nl
Similarly, we do not have access to an ideal $K$. Instead, we have to estimate it using data and inductive modeling assumptions \citep[e.g. ][]{rothfuss2019noise}. Ultimately, our model and property descriptions may be evaluated using not just the training and test dataset $\set{D}$ (see \Cref{subsec:WhatMLmodelsRepresent}), but also relevant available unlabeled data or artificially generated data. We call this bundle the \emph{evaluation data} $\set{D}^*\supseteq \set{D}$.
\nl
A way to estimate property descriptions with access only to the ML model and evaluation data is provided by property description \emph{estimators}, which we assume to be unbiased estimators of $g_K$, i.e. the function $\hat{g}_{\set{D}^*}:\set{M}\rightarrow \set{Q}$ fulfills
\begin{equation*}
    \mathbb{E}_{\rv{D}^*} [\hat{g}_{\rv{D}^*}(\hat{m})]=g_K(\hat{m}) \quad \text{for all }\hat{m}\in\set{M}.
\end{equation*}

\begin{figure}[h]
\centering

     \begin{subfigure}[b]{0.4\textwidth}
         \centering
         \includegraphics[width=\textwidth]{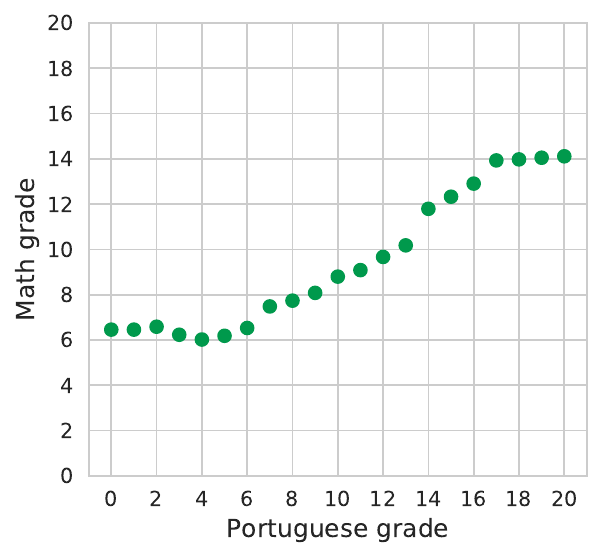}
         \caption{cPDP}
         \label{fig:cpdp}
     \end{subfigure}
    \begin{subfigure}[b]{0.411\textwidth}
         \centering
         \includegraphics[width=\textwidth]{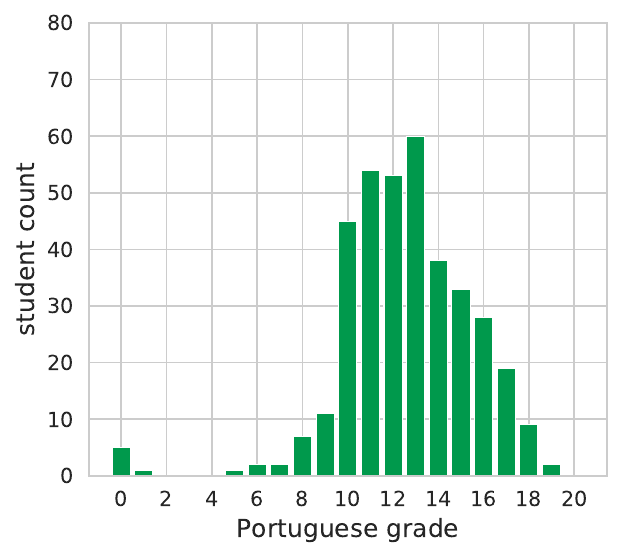}
        \caption{histogram}
        \label{fig:histogram}
     \end{subfigure}
     \caption{(a) shows the cPDP estimate of $\mathbb{E}_{\rv{Y}\cond \rv{X}_p}[\rv{Y}\cond\rv{X}_p]$ via \Cref{eq:approxDescriptor}. Note that the grade jittering strategy described in the text allows us to evaluate e.g. $\hat{m}(x_p=3)$ even though we have no data for that value. (b) shows the histogram of grades in Portuguese in the original dataset \citep{cortez2008using}.}
\end{figure}

\paragraph{Example: practical estimates of property descriptions}
Our evaluation dataset $\set{D}^*$ comprises the initial training and test dataset $\set{D}$ as well as artificial instances created by the following manipulation: we make six full copies of the data, and jitter the Portuguese grade by  $1,-1,2,-2,3$ or $-3$ respectively in each of them. This augmentation strategy reflects how we understand the Portuguese grade as noisy based on our background knowledge of how much student performance varies daily and teachers grade inconsistently. Let the students with jittered Portuguese grade $i$ be $\set{D}^*_{|x_p=i}\coloneqq (x\in \set{D}^* \cond x_p=i)$, then, we can calculate the property description estimator at $i$ as the following conditional mean (an unbiased estimator of the conditional expectation):
\begin{equation}
    \hat{g}_{\set{D}^*}(\hat{m})(i)\coloneqq \frac{1}{|\set{D}^*_{|x_p{=}i}|}\sum\limits_{x\in \set{D}^*_{|x_p=i}}\hat{m}(x).
    \label{eq:approxDescriptor}
\end{equation}
The estimated answer to our initial question is plotted in \Cref{fig:cpdp}. Math grades appear to depend on Portuguese grades only in the range $x_p\in(8\text{--}17)$. However, \Cref{fig:histogram} shows that we have very sparse data in some regions (e.g. very few students scored below $x_p=8$), a fact that we must take into account before confirming this first impression.

\subsection*{Step 4: Quantify the uncertainties in property descriptions}
We have shown how we can estimate $Q$ using an approximate ML model paired with a suitable evaluation dataset. But how good is our estimate? 
\nl
In estimating $Q$ using the ML model $\hat{m}$ instead of the optimal model $m$, we introduce a \emph{model error}, $\mathrm{ME}[\hat{m}]:=d_{\set{Q}}\bigl(g_K(m),g_K(\hat{m})\bigr)$. Moreover, by using the evaluation data $\set{D}^*$ instead of knowledge $K$, we also introduce an \emph{estimation error}, $\mathrm{EE}[\set{D}^*]:=d_\set{Q}\bigl(g_K(\hat{m}),\hat{g}_{\set{D}^*}(\hat{m})\bigr)$.
\nl
In theory, the model error and the estimation error can be cleanly separated. In practice, however, they are often statistically dependent because the training and the evaluation data overlap. Fortunately, there exist various sample splitting approaches that allow to circumvent or minimize this dependence \citep{james2023resampling}. Generally, neither the model error nor the estimation error can be computed perfectly; this would require access to the optimal model $m$ and infinitely many data instances. Nevertheless, we can quantify them in expectation.
\nl
An intuitive approach to quantifying the expected errors is to decompose them into bias and variance contributions. For the bias-variance decomposition, we assume the metric $d_\set{Q}$ to be the squared error.\footnote{A bias-variance decomposition is also possible for other loss functions, including the 0-1 loss \citep{domingos2000unified}.} Considering the dataset that we entered into the learning algorithm as a random variable $\rv{D}$, we can decompose the expected model error as follows
\begin{equation*}
    \mathbb{E}_{\rv{D}}[\mathrm{ME}[\hat{m}]]=\underbrace{(g_K(m)-\mathbb{E}_{\rv{D}}[g_K(\hat{m})])^2}_{\mathrm{Bias}^2}\;+\;\underbrace{\mathbb{V}_{\rv{D}}[g_K(\hat{m})]}_{\mathrm{Variance}},
\end{equation*}
where $\hat{m}\coloneqq I(\set{D})$ is the trained model (output of a machine learning algorithm $I$ for dataset $\set{D}$, \Cref{subsec:WhatMLmodelsRepresent}). Similarly, considering the evaluation data as a random variable $\rv{D}^*$, we can also decompose the expected estimation error as follows
\begin{equation*}
    \mathbb{E}_{\rv{D}^*}[\mathrm{EE}[D^*]]=\underbrace{(g_K(\hat{m})-\mathbb{E}_{\rv{D}^*}[\hat{g}_{\rv{D}^*}(\hat{m})])^2}_{\mathrm{Bias}^2}\;+\;\underbrace{\mathbb{V}_{\rv{D}^*}[\hat{g}_{\rv{D}^*}(\hat{m})]}_{\mathrm{Variance}}=\mathbb{V}_{\rv{D}^*}[\hat{g}_{\rv{D}^*}(\hat{m})].
\end{equation*}
In this case, the bias term vanishes because the property description estimator is by definition unbiased w.r.t. the property descriptor.
\nl
There are indeed different approaches to quantify the uncertainties of property descriptions. The standard frequentist approach is to estimate the variance in above equations using refitted models and property descriptions with resampled data \citep{Molnar2023relating}. But there are also Bayesian approaches where uncertainty is directly baked into the prediction model: BART by \cite{chipman2012bart}, for example, uses a sum-of-trees to perform Bayesian inference and directly provides uncertainties for inferential quantities like marginal effects. Similarly, Gaussian processes provide predictive error-bars \citep{rasmussen2010gaussian}, which offer a natural confidence measure for property descriptions  \citep{moosbauer2021explaining}. Finally, Bayesian posteriors can even be obtained for neural network architectures \citep{gal2016theoretically,van2020uncertainty} and leveraged to estimate the uncertainty of property descriptions like counterfactuals \citep{holtgen2021deduce,schut2021generating}.

\begin{figure}[h]
\centering

     \begin{subfigure}[b]{0.4\textwidth}
         \centering
         \includegraphics[width=\textwidth]{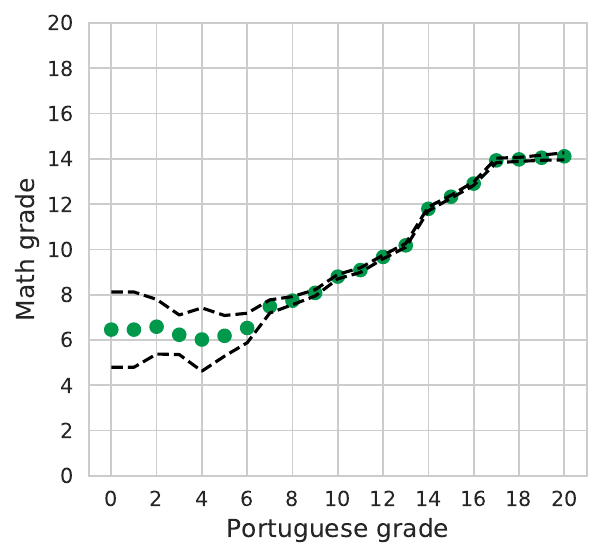}
         \caption{estimation error}
     \end{subfigure}
    \begin{subfigure}[b]{0.4\textwidth}
         \centering
         \includegraphics[width=\textwidth]{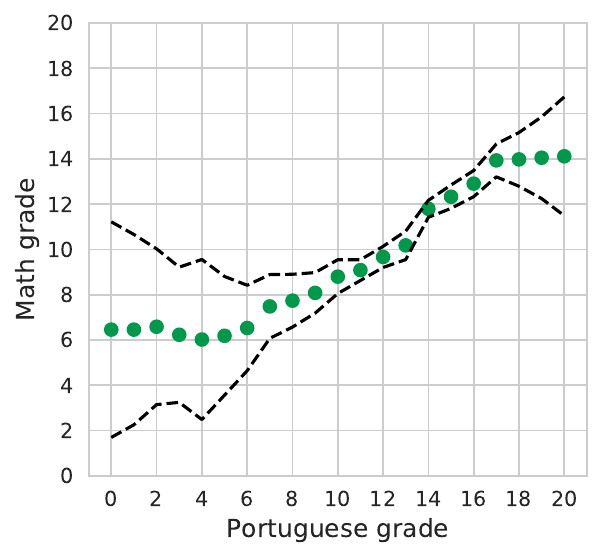}
        \caption{estimation + model error}
     \end{subfigure}
     \caption{(a) shows the cPDP (dots) and its estimation error due to Monte-Carlo integration (dashed lines, Eq.~\ref{eq:conf1}). (b) shows the cPDP with \emph{both} estimation and model error (Eq.~\ref{eq:conf2}). Confidence bands cover the true expected math grade in 95\% of all cases. These plots jointly suggest that most of the uncertainty is due to model error.}
     \label{fig:cpdpConfidence}
\end{figure}

\paragraph{Example: uncertainty in property descriptions}
We will certainly obtain different cPDPs (\Cref{fig:cpdp}) if we use different models with similar performance, or different subsets of evaluation data, so how much can we then rely on these cPDPs?
\nl
The estimates of the variances of the cPDP by \cite{Molnar2023relating} allow us to calculate pointwise confidence intervals (\Cref{fig:cpdpConfidence}). We can calculate a confidence interval at significance $\alpha$ that only incorporates the estimation uncertainty due to finite data as follows:
\begin{equation}
    \mathrm{CI}_{\mathrm{EE}[\rv{D}^*]}\coloneqq \hat{g}_{\set{D}^*}(\hat{m})(i)\pm t_{1-\frac{\alpha}{2}} \sqrt{\hat{\mathbb{V}}_{\rv{D}^*}[\hat{g}_{\rv{D}^*}(\hat{m})(i)]},
    \label{eq:conf1}
\end{equation}
where $t_{1-\alpha/2}$ is the critical value of the $t$-statistic. We can also obtain a confidence interval that incorporates both model and estimation uncertainty:
\begin{equation}
    \mathrm{CI}_{\mathrm{ME}[\hat{m}]\land\mathrm{EE}[\rv{D}^*]}\coloneqq \hat{g}_{\set{D}^*}(\hat{m})(i)\pm t_{1-\frac{\alpha}{2}} \sqrt{\hat{\mathbb{V}}_{\rv{D},\rv{D}^*}[\hat{g}_{\rv{D}^*}(\hat{m})(i)]}.
    \label{eq:conf2}
\end{equation}
For this combined confidence interval to be valid, the estimation of the property descriptions must be unbiased. While unbiasedness of the property description estimator and the ML algorithm\footnote{Unbiasedness of the algorithm means that, in expectation over training sets, the ML algorithm learns the optimal model, i.e. $m=\mathbb{E}_{\rv{D}}[\hat{m}]$. Since unbiasedness is context-specific, there is no conflict with the no-free-lunch theorems \citep{sterkenburg2021no}.}
is sometimes sufficient to prove the unbiaseness of property descriptors \citep{Molnar2023relating}, there often is a tension between the bias-variance trade-off made to obtain global model fit versus the best estimate of the more localized property descriptions \citep{van2011targeted}. Fortunately, there exist various debiasing strategies using influence functions \citep{hines2022demystifying} like the targeted maximum likelihood update \citep{van2006targeted} or orthogonalization \citep{chernozhukov2018double}.
\nl
\Cref{fig:cpdpConfidence} shows that for students with Portuguese grades between $8$ and $17$, we can be very confident in our model and the relationship it identifies between math and Portuguese grade.\footnote{We used bootstrapping to estimate the two variances. In non-synthetic-data settings, it is generally not possible to always sample new data for the model training and the evaluation. Although bootstrapping may underestimate the variance, our goal here is simply to illustrate the process of quantifying uncertainty for a concrete IML method.} However, for Portuguese grades outside this range, the true value might be far off from our estimated value, as we can see from the width of the confidence intervals. For these grade ranges, gathering more data may reduce our uncertainty.

\subsection*{Synopsis of the steps} 
\begin{figure}[h]
\centering
   \includegraphics[width=0.8\linewidth]{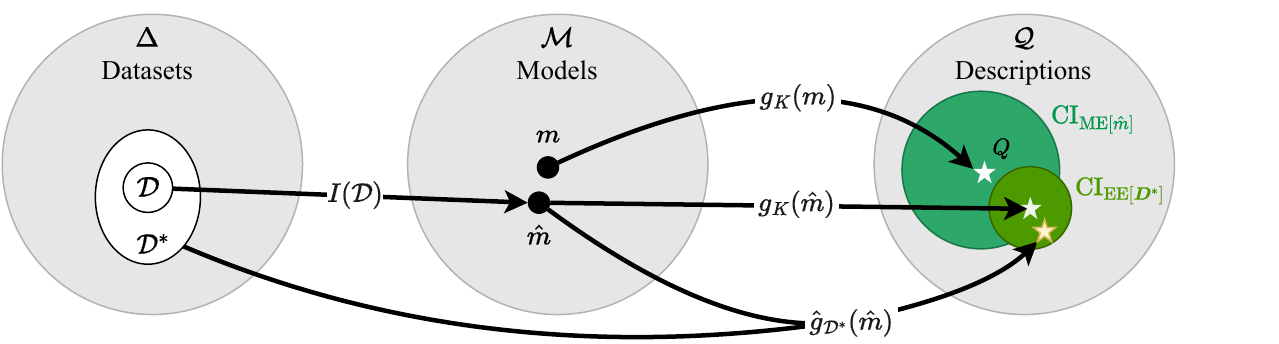}
  \caption{\textbf{From datasets to inferences via ML models}. Mappings (arrows) connect datasets with models and descriptions (points within sets represented as ovals; confidence intervals in shades of green). Practical IML descriptions $\hat{g}_{\mathcal{D}^\star}(\hat{m})$ (bottom arrow) approximate with quantifiable uncertainty an estimand $Q$ by using a model $\hat{m}$ fit on $\mathcal{D}$ together with an evaluation dataset $\mathcal{D}^*$.} 
\label{fig:commDiagram}
\end{figure}

We provide in \Cref{fig:commDiagram} an overview of the functions and spaces involved in IML for scientific inference. We started from a phenomenon and formalized a scientific question---our estimand $Q$. Using a learning algorithm $I$ on a dataset $\set{D}$ representative of the phenomenon, we trained an ML model $\hat{m}$ that approximates the optimal model $m$. We then set out to estimate $Q$ from $\hat{m}$. We defined a property descriptor $g_K$, that is, a function that allows to compute $Q$ from $m$ given $K$, respectively approximates $Q$ from $\hat{m}$ given $K$. Because $g_K$ requires probabilistic knowledge about $\prob(\rvs{X},\rv{Y})$, which can only be obtained from data, we introduced a property description estimator $\hat{g}_{\set{D}^*}$---a function estimating $Q$ solely from a finite evaluation dataset $\set{D}^*$. Finally, we showed how the expected error due to our approximate modeling and finite-data-estimation can be quantified with respective confidence intervals $\mathrm{CI}_{\mathrm{ME}[\hat{m}]}$ and $\mathrm{CI}_{\mathrm{EE}[\rv{D}^*]}$.

\section{Some IML methods already allow scientific inference}
\label{sec:commonlyInterestingPhenomenonProp}

Many estimands are relevant across a wide variety of scientific domains. The goal of practical IML research for inference should be to define practical property descriptors for such estimands and provide accessible implementations of these descriptors, including quantification of uncertainty. To find out about scientifically relevant estimands, IML researchers, statisticians, and scientists must interact closely. 

\begin{table}
\SetTblrStyle{caption-tag}{font=\bfseries}
\begin{talltblr}[
label={tab:CommonlyRelevantDes}, 
caption={\textbf{Global and local questions with their matching estimands and property descriptors}. Row background indicates type of question: white for global, gray for local ones. The set $S$ in $\rvs{X}_{S}$ is a subset of the total feature set $S\subseteq\mathcal{N}:=\{1,\dotsc,n\}$; $\rvs{X}$ and $\rvs{X}_{-p}$ abbreviate $\rvs{X}_\mathcal{N}$ and $\rvs{X}_{\mathcal{N}\setminus \{p\}}$ respectively, as in the rest of the manuscript. The model $m_{\rvs{X}_S}$ is the optimal predictor of $\rv{Y}$ w.r.t. loss $L$ and features $S$. The support of $\rvs{X}$ is denoted $\operatorname{supp}\ \rvs{X}:=\bm{x}\in\set{X} \land \prob(\rvs{X}=\bm{x})>0$. See text for other notations.}, 
note{a} = {\footnotesize PRIM is just one approach to maximum a posteriori inference \citep[Chapter 5]{murphy2022probabilistic}.}, 
note{b} = {\footnotesize Only with the right similarity metric that accounts for the plausibility constraint.}]{
  width = 1\linewidth, colspec = {X[1.2,l] X[3,c] X[4.2,c] X[1.4,c]},
  cell{1}{1} = {c=2}{c}, 
  row{1} = {font=\bfseries, gray!40, abovesep=6pt, belowsep=6pt}, 
  row{4,5,8,9,12,13,16,17} = {gray!20}, 
  cell{2,6,10,14}{1} = {r=4}{halign=l, font=\bfseries}, 
  cell{even[2]}{2,3} = {r=2}{c}, 
  column{1} = {colsep=1pt}, 
  column{3} = {colsep=2pt}, 
  cell{2-Z}{3} = {mode=math}, 
  cell{odd[2]}{4} = {font=\tiny}, 
  row{even[2]} = {abovesep=4pt}, 
  vline{2-4} = {1}{dotted, 1pt},
  hline{odd} = {0}{dash=dashed},
  hline{even} = {-}{} 
}
\toprule
Global / local question & & Estimand & IML method\\
\midrule
Effect & What is the best estimate of $\rv{Y}$ if we only know $\rv{X}_p$? 
       & m_{\rv{X}_p}(\rv{X}_p) & cPDP \\ & & & \cite{apley2020visualizing} \\
       & How does the best estimate of $\rv{Y}$ change relative to $\rv{X}_p$, knowing that $\rvs{X}_{-p}=\bm{x}_{-p}$? 
       & m_{\rvs{X}}(\rv{X}_p,\bm{x}_{-p}) & ICE curve \\ & & & \cite{goldstein2015peeking} \\
Conditional contribution & How much worse can $\rv{Y}$ be predicted from $\rvs{X}$ if we did not know $\rv{X}_p$? 
       & {\rm EPE}_{\rvs{X}_{-p},\rv{Y}}\ m_{\rvs{X}_{-p}}(\rvs{X}_{-p}) - {\rm EPE}_{\rvs{X},\rv{Y}}\ m_{\rvs{X}}(\rvs{X}) & cFI \\ & & & \cite{strobl2008conditional} \\
       & How much worse can $\rv{Y}$ be predicted from $\rvs{X}=\bm{x}$ if we did not know $\rv{X}_p$? 
       & L(y,m_{\rvs{X}_{-p}}(\bm{x}_{-p})) - L(y,m_{\rvs{X}}(\bm{x})) & ICI \\ & & &  \cite{casalicchio2019visualizing} \\
{Average \\\hspace*{0pt}contribution} & What is on average the share of feature $\rv{X}_p$ in the prediction of $\rv{Y}$ across feature coalitions? 
        & \underset{S{\subseteq}\mathcal{N}{\setminus}\{p\}}{\frac{1}{n}{\displaystyle \sum}}\binom{n-1}{|S|}^{-1} \Bigl({\rm EPE}_{\rvs{X}_{S},\rv{Y}}\ m_{\rvs{X}_{S}}(\rvs{X}_{S}) - {\rm EPE}_{\rvs{X}_{S{\cup}\{p\}},\rv{Y}}\ m_{\rvs{X}_{S{\cup}\{p\}}}(\rvs{X}_{S{\cup}\{p\}})\Bigr)
       & SAGE \\ & & & \cite{covert2020understanding} \\
       & What is on average the share of feature $\rv{X}_p$ in the prediction of $\rv{Y}$ if $\rv{X}=\bm{x}$ across feature coalitions? 
       & \underset{S{\subseteq}\mathcal{N}{\setminus}\{p\}}{\frac{1}{n}{\displaystyle \sum}}\binom{n-1}{|S|}^{-1} \bigl(m_{\rvs{X}_{S}}(\bm{x}_S) - m_{\rvs{X}_{S{\cup}\{p\}}}(\bm{x}_S,x_p)\bigr)
       & cSV\\ & & &  \cite{aas2021explaining} \\
Relevant value & Under which realistic conditions $\rvs{X}$ can we observe relevant value $y_{\rm rel}$? 
       & \underset{\bm{x}\in\operatorname{supp} \rvs{X}}{\argmin} d_\set{Y}(m_{\rvs{X}}(\bm{x}),y_{\rm rel}) & PRIM\TblrNote{a}  \\ & & & \cite{friedman1999bump} \\
      & Under which realistic conditions similar to $\bm{x}$ can we observe relevant value $y_{\rm rel}$? & \underset{\bm{x}'\in \operatorname{supp} \rvs{X}}{\argmin} d_{\set{Y}}(m_{\rvs{X}}(\bm{x}'),y_{\rm rel})+\lambda\; d_{\set{X}}(\bm{x},\bm{x}')  & \hspace*{0pt}Counterfactuals
       \TblrNote{b} \\ & & & \cite{dandl2020multi}  \\
       \bottomrule 
\end{talltblr}
\end{table}

In \Cref{tab:CommonlyRelevantDes} we present a few examples of practical inference questions that can be addressed by existing IML methods, i.e., these methods can operate as property descriptors already. We distinguish between global and local questions about the phenomenon: global questions concern general associations (e.g.between math and Portuguese grade), local questions concern associations in the local neighborhood of an instance (e.g. between study time and math grade for a specific student), and appear with gray background. The last column identifies current IML methods that provide approximate answers, albeit often without uncertainty quantification. To draw scientific inferences, we ultimately need versions of IML methods that account for the dependencies in the data \citep{hooker2021unrestricted}.
\nl
Not only the IML literature has worked on property descriptors, but the fairness literature also discussed measures that can be described as property descriptors, e.g. statistical parity (see \cite{verma2018fairness} for an overview). Similarly, robustness measures under distribution shifts (see \cite{freiesleben2023beyond} for an overview) as well as methods that examine the system stability in physics-informed neural networks \citep{chen2018neural,raissi2019physics} can be seen as property descriptors.

\subsection*{Example: illustrating the methods from Table 2 on the grading example} 
To illustrate how the IML methods from \Cref{tab:CommonlyRelevantDes} can help to address general inferential questions, we will introduce them along our grading example. We will begin with a discussion of global questions before moving on to local questions.

\begin{figure}[h]
\centering
   \includegraphics[width=0.8\linewidth]{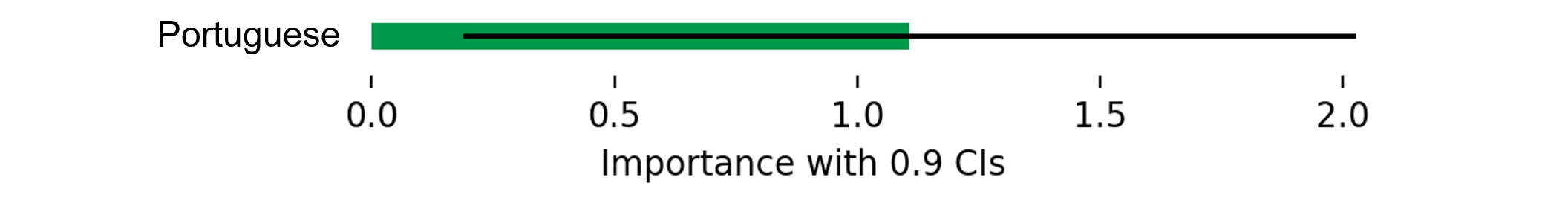}
  \caption{\textbf{Conditional feature importance of the Portuguese grade with 90\% confidence interval}. The confidence interval, which takes into account the model and estimation uncertainty, does not contain the value zero. On this basis, a researcher may reject the null hypothesis that the Portuguese grade is conditionally independent of the math grade given the remaining features.}
\label{fig:cFI}
\end{figure}

\paragraph{cFI} 
We have seen in the pedagogical example above that the association between language and math skills can be inferred using the cPDP. Another question is whether language skill provides information about math skill that is not contained in other student features (e.g., study-time, absences, and parents educational background). This is a common question among education scientists \citep{peng2020examining} and can be formalized by asking if language skill $\rv{X}_p$ is independent of math skill $\rv{Y}$, conditional on the remaining features $\rvs{X}_{-p}$:
\[H_0: \rv{X}_p \perp \rv{Y} | \rvs{X}_{-p}.\]
To test conditional independence, conditional feature importance (cFI) can be used \citep{strobl2008conditional,konig2021relative,watson2021testing,ewald2024guide}. cFI compares the performance of the model before and after perturbing the feature of interest while preserving the dependencies with the remaining features. If the Portuguese grade is conditionally independent of the math grade, all relevant information in the Portuguese grade can be reconstructed from the remaining features so that the performance is not affected by the perturbation. Thus, if the cFI is nonzero, Portuguese grade must be conditionally dependent with the math grade. To account for the uncertainties, we rely on \cite{Molnar2023relating}.
\nl
In \Cref{fig:cFI}, we applied cFI to our ML model $\hat{m}$ and computed the respective confidence interval (quantifying both model and estimation uncertainty). The importance of the Portuguese grade for the math grade is significant according to the 90\% confidence intervals, as zero is not contained in the interval. On this basis, the scientist may reject $H_0$.

\paragraph{SAGE} 
Similar inferential questions as with the cFI can be addressed with Shapley additive explanations \citep[SAGE,][]{covert2020understanding}. In contrast to the cFI, however, SAGE averages conditional importance relative to each subset of the features, often referred to as coalitions. Even if a feature has a cFI of zero, the corresponding SAGE values can be positive because the feature has positive importance in at least one of the coalitions \citep{ewald2024guide}.
\nl
SAGE values are based on so-called surplus contributions of a feature of interest $p$ (say the Portuguese grade) with respect to a coalition $S$ (e.g., the number of absences and neighborhood). More specifically, surplus contributions quantify how model performance changes when the model, which only has access to features in the coalition $S$, additionally gets access to $j$. \footnote{In contrast to cFI, where features are removed from the model via perturbations, SAGE removes variables from the model by marginalizing them out using the conditional expectation.} 
\nl
Notably, the surplus of a feature depends on the coalition $S$: Adding a dependent feature to the coalition decreases the surplus, e.g. the surplus of the current Portuguese grade is lower if we already know the Portuguese grade from last term. Conversely, adding a collaborating feature can increase a feature's surplus, e.g., the effect of the mother being unemployed may be larger if the father is also unemployed. By averaging over the surplus of a feature across all possible coalitions (weighted by the number of possible feature orderings in which the coalition precedes the feature of interest), SAGE provides a broad insight into the importance of a feature. The idea behind SAGE is based on the Shapley value \citep{covert2020understanding}, a concept from cooperative game theory \citep{shapley1953value}.

\paragraph{PRIM} What defines the optimal math student? Education scientists have clear ideas about general indicators of student success in math, such as parents' social and economic status, students' habits and motivation, and cultural factors \citep{kuh2006matters}.  On this basis, they may construct a hypothetically optimal math student: highly educated parents who work in STEM professions, high study time and frustration tolerance, and a cultural environment that values education. One approach to test scientists' intuitions about the optimal math student is to compare her expected success with the expected success of the optimal student(s) according to the data distribution. Patient rule induction \citep[PRIM,][]{friedman1999bump} estimates the student(s) with optimal conditions according to the data distribution, thus allowing scientists to test their hypotheses. Note however that the evaluation of such a test can be difficult: The optimal math student(s) according to scientists may lie outside of the data distribution, making their expected success difficult to estimate. In general, the estimation of high-dimensional feature vectors rather than scalars may be statistically less stable, leading to greater uncertainty.

\paragraph{ICE} Education scientists may wish to infer how study time statistically influences expected student success of individual students in math \citep{rohrer2007increasing}. One way to formalize this question is by the conditional expectation of the math grade given a set of input features where study time is varied. This is exactly the quantity that individual conditional effect curves \citep[ICE,][]{goldstein2015peeking} estimate. However, caution is advised, as the variation of study time may break dependencies with other features, forcing the model to extrapolate \citep{hooker2021unrestricted}. On the basis of ICE curves, education scientists may hypothesize that for a subset of students, namely efficient learners, the effect of increasing study time on students' performance saturates
\citep{rohrer2007increasing}; a hypothesis that can then be tested in a different study.

\paragraph{cSV and ICI} Education scientists want to understand the reasons behind the low expected success of some specific students in math \citep{saha2024factors}. One way to approach this question is to investigate how knowing certain features (e.g. study time, language grades, or absences) affects the expected math grade for specific low performing students. One property descriptor that allows to approach this question is the conditional Shapley Value \citep[cSV,][]{aas2021explaining}.
\nl
Like SAGE values, cSV are Shapley value methods based on averaging the surplus contribution of a feature $j$ across all possible coalitions $S$; In contrast to SAGE surplus contributions, cSV surplus contributions do not quantify the surplus in prediction performance for all students, but the change in the predicted value for a specific student. In our example, the cSV of the Portuguese grade $p$ for an individual $x$ tells us how the expected value of the student's math grade changes when we get access to the Portuguese grade $x_p$, averaged over all possible coalitions $x_S$ of features that we already know.
\nl
Individual conditional importance \citep[ICI,][]{casalicchio2019visualizing} also estimates the importance of individual features for the expected math grade of a given student, but not the effect on the prediction itself, but on the accuracy of the prediction (in terms of its loss). 

\paragraph{Counterfactuals} Education scientists are interested in identifying feasible conditions that increase the expected success in math of individual students \citep{saha2024factors}. For a given student, this question can be framed as a search for similar alternative student features with higher expected math grades. This is the target estimand of plausible counterfactual explanations \citep{dandl2020multi}. Analyzing the difference between the original instance and the alternative student features allows education scientists to identify features that are locally associated with higher student success. For example, plausible counterfactuals for a specific student with a bad math grade, $6$ absences and a mediocre Portuguese grade suggest that a reduced number of absences to $3$ would have yielded a better expected math grade. Counterfactuals do only reflect associations in the data and should not be interpreted causally (see \Cref{sec:causality}), however, based on counterfactuals, scientists can derive hypotheses about causally relevant factors, which may then be tested in an experimental study. However, it should be noted that there can be (infinitely) large numbers of such counterfactual instances, and selecting among them requires domain knowledge \citep{mothilal2020explaining}.

\subsection*{The delicate line between exploration and inference}
Some of the examples just presented, particularly regarding local descriptors, are concerned with exploring data properties rather than drawing concrete inferences using statistical tests. A delicate distinction, according to which inference requires a constrained set of hypotheses to be tested, whereas exploration describes the search for hypotheses \citep{tredennick2021practical}.\footnote{The motivation for the distinction between inference and exploration is the multiple comparisons problem \citep{tredennick2021practical} -- if many hypotheses are tested simultaneously, the probability that some inferences will be false discoveries is very high. Note that there are further approaches to tackle the multiple-comparison problem, such as the Bonferroni correction \citep{curran2000multiple,lindquist2015zen}.} Our notion of inference -- investigating unobserved variables and parameters to draw conclusions from data -- is broader and encompasses data exploration. We motivated local property descriptors with these rather exploratory questions because the data support for local estimands is very limited, which leads to greater uncertainty and, consequently, to inferential tests with low statistical power. For this reason, local descriptors are rarely used in practice to test concrete hypotheses.

\subsection*{Disagreement between descriptors}
There is a growing concern in the IML community about the fact that different IML methods disagree. For example, disagreement has been demonstrated for common feature attribution techniques such as Shapley values and LIME, but also for commonly used saliency maps \citep{krishna2022disagreement,adebayo2018sanity,ghorbani2019interpretation}. While at first glance this may indicate that the methods have limited use in science, they can only diverge meaningfully if they address different inferential questions. Given two property descriptors have the same estimand, they must by definition (given enough data) converge to the same property descriptions.

\section{Property descriptors do not generally provide causal inferences}
\label{sec:causality}
With property descriptors, we can access a wealth of properties of the observational joint distribution that answer various scientific questions. While the observational joint probability distribution is indeed interesting, it remains on rung 1 of the so-called ladder of causation---the associational level \citep{pearl2018book}. What scientists are often much more interested in, is answering causal questions, such as what is the average treatment effect (rung 2) \citep{imbens2015causal} or even counterfactual, what-if questions \citep[rung 3;][]{salmon1998causality,sep-scientific-explanation}. In our example, we may be interested not only in how strongly students' language and math skills are associated (rung 1), but also in how much the provision of tutoring in Portuguese affects students' math skills (rung 2) or whether a specific student (who is not a native Portuguese speaker) would have done better in math had she received Portuguese tutoring at a young age (rung 3).
\nl
Supervised ML models only represent aspects of the observational distribution (rung 1)  and, therefore, do not directly provide causal insights \citep{pearl2019limitations}.
Consequently, property descriptions do not provide causal insights either.
Many IML works that discuss causality \citep{schwab2019cxplain,janzing2020feature,wang2021shapley} are only concerned with causal effects on model \emph{predictions}, which do not necessarily translate into a causal insight into the phenomenon \citep{konig2023improvement}---they address only model audit.
\nl
Machine learning methods can still be used to gain causal insights into natural phenomena, however, only if additional causal assumptions are posed. For example, if the so-called  backdoor criterion is fulfilled in the causal graph, we can identify the average causal effect from observational data using the backdoor adjustment formula \citep{pearl2009causality}. Or, formulated in terms of the Potential Outcomes (PO) framework
\footnote{Pearl's causal modeling framework and Rubin's potential outcomes framework both concern causal inference and are logically equivalent -- which one to choose depends on personal preference and the practical use case \citep{pearl2010causal,imbens2020potential}.}
by \cite{rubin1974estimating}: if conditional exchangeability is fulfilled, we can use the adjustment formula. Given the causal effect can be identified, there are various ML-based approaches to estimate it, like the T-learner, the S-learner, and doubly robust methods (see \cite{kunzel2019metalearners,knaus2022double} or \cite{dandl2023causality} for an overview). Prominently, there is double ML \citep{chernozhukov2018double,fink2023double} and targeted learning \citep{van2011targeted} that provide unbiased estimates of various identifiable causal estimands. Note, however, that to arrive at the necessary knowledge, we require interventional data and/or have to make strong, untestable assumptions \citep{rubin1974estimating,holland1986statistics,meek2013strong,spirtes2000causation}.
Further, even if a causal estimand is identifiable and can therefore be estimated with ML, estimation from finite data may be challenging \citep{kunzel2019metalearners}.
\nl
In certain simplified scenarios, IML methods applied to associational ML models can be helpful for causal inference. Firstly, if all predictor variables are causally independent and the features cause the prediction target, the causal model interpretation implies the causal phenomenon interpretation. Secondly, associative models in combination with IML can help estimate causal effects even in the absence of causal independence if they are, in principle, identifiable by observation. For example, the partial dependence plot coincides with the so-called adjustment formula; It, therefore, identifies a causal effect if the backdoor criterion (conditional exchangeability) is met and the model optimally predicts the conditional expectation \citep{zhao2021causal}.  Thirdly, when there is access to observational and interventional data during training, ML models trained with invariant risk minimization predict accurately in interventional environments \citep{peters2016causal,pfister2021stabilizing,arjovsky2019invariant}. For such intervention-stable models, IML methods that provide insight into the effect of interventions on the prediction also describe causal effects on the underlying real-world components \citep{konig2023improvement}.
\nl
While supervised learning learns from a fixed dataset, reinforcement learning (RL) systems are designed to act and can, therefore, assess the effect of their interventions. As such, RL models can be designed to provide causal interpretations \citep{bareinboim2015bandits,zhang2017transfer,gasse2021causal}.
\nl
Finally, another way in which ML supports causal inference is by facilitating practical scientific inference based on potentially complex, but still ER, mechanistic models that are frequently implemented as numerical simulators. Indeed, simulators can represent complex, causal dynamics in an ER fashion, but often at the price of an intractable likelihood and, thus, expensive or even intractable inference. A variety of new methods for likelihood-free inference \citep{cranmer2020frontier} allows us to estimate a full posterior distribution over ER parameters for increasingly complex models using ML.

\section{Discussion}
\label{sec:discussion}
ER models enable straightforward scientific inference because their elements are meaningful: they directly represent elements of the underlying phenomenon. While ML models are generally not ER, property descriptors can offer an indirect route to scientific inference, provided whole-model properties have a corresponding phenomenon counterpart. We have shown how phenomenon representation can be accessed through optimal predictors and described how to practically construct property descriptors following four steps: the first two steps clarify how domain questions and property descriptors can be theoretically connected, step three shows how to practically estimate property descriptions with ML models and data, and step four allows to evaluate how much the estimated property descriptions may deviate from the ground truth. We highlighted some current IML methods that can already be seen as property descriptors and identified what inference questions they answer.

\subsection*{Scope for HR modeling in science}
ML models represent just an extreme case where almost no element of the model is representational. There is a continuum between ER models and HR-only models---ranging from full ER models like Newton's gravitational laws to intermediate statistical models containing higher-order interaction terms to full-blown HR models like deep neural networks. Our main message is: the four-step approach can be used to extend inference to any non ER model, whether ML or not.
\nl
Why should scientists use HR models at all? \cite{rudin2019stop} argued that whenever there is an HR model with high predictive accuracy, there is also an ER model that achieves similar performance. She backs up her argument with several examples, e.g. loan risk prediction \citep{rudin2019we}, recidivism prediction \cite{zeng2017interpretable}, finding pattern in EEG data \citep{li2017targeting}, and even image classification \citep{chen2019looks}, where (relatively) interpretable models achieve performance comparable to less interpretable HR models. She therefore recommends favoring ER models in high-stakes settings. 
\nl
In science, the stakes are quite high and interpretability is highly valued. Therefore, we believe that highly-predictive ER models, when available, should be favored in science over HR models. The problem is that while such powerful ER models may exist, they can be hard to find, especially in model classes with high complexity \citep{nearing2021role}. While reducing the complexity of the model class can help to find high-performing ER models in noisy environments \citep{semenova2024path}, this requires substantial domain knowledge about the data generating process, which is often not available. 
Choosing simplified models with little predictive performance just for the sake of inherent interpretability is no viable path \citep{breiman2001statistical}. Interpreting models that poorly approximate the phenomenon will lead to unreliable conclusions \citep{cox2006principles,good2012common}.\footnote{Note however that even for the optimal model, there remains the so-called Bayes error rate, an irreducible error arising from the fact that $\rvs{X}$ does not completely determine $\rv{Y}$ \citep{hastie2009elements}. Thus, high error does not necessarily flag a low-quality model, but rather may indicate that $\rvs{X}$ provides insufficient information about $\rv{Y}$.} 
 \nl
 However, there could be a reasonable compromise between ER and HR modeling, where some parts of the model are ER, while other parts, where less parametric assumptions are justified, are filled with powerful HR models: This approach is commonly taken in the intersection between causal modeling and ML, where the causal graph is ER whereas the functional dependencies are modeled with HR models \citep{peters2017elements}. Similarly, using flexible ML models while enforcing constraints in the training process like additivity can lead to partially ER models \citep{hothorn2010model,van2011targeted}. In physics we may want to model a phenomenon with a classical ER model, namely ordinary differential equations, without constraining the function \citep{dai2022kernel}, or we may want to enforce a preference for certain functions (e.g. exponential or sinus) without limiting the possible dependencies \citep{martius2016extrapolation}.

\subsection*{Limitations of our framework}
\label{subsec:openProb}

\paragraph{How useful are property descriptors for scientific inference in practice?}
We have shown that, under certain assumptions, property descriptors provide insight into the phenomenon. However, we have not shown that property descriptors are the best approach to gain this insight. The cPDP, for example, could also be estimated directly from the data without interposing an ML model and a property descriptor. Does it make sense to take the detour via the ML model and property descriptors instead of directly estimating the quantity of interest, e.g. using targeted learning \citep{van2011targeted}? 
\nl
One case is when access to the model, the data, or computational resources is limited, which is common with proprietary models like ChatGPT or other sophisticated models like Alphafold \citep{senior2020improved} or ResNet \citep{he2016deep}. Due to the data and expertise that went into these models, they are ideal candidates for mining them for insights with IML tools. However, computing interpretations for these models can again be computationally very expensive. Another use case of property descriptors is when scientists have set out to obtain predictions but want to gain additional insights from their model at low cost. 
\nl
Irrespective of whether scientists should use property descriptors to make concrete scientific inferences, ample evidence in the published scientific literature shows that scientists use IML tools and draw inferences based on these interpretations \citep{roscher2020explainable,shahhosseini2020forecasting,gibson2021training}. Our paper can help clarifying what inferences scientists are enabled to draw from interpretations and what IML tools to use.
\nl
To make a fair comparison between the inferential qualities of targeted learning and property descriptors a systematic study would be needed. Under what conditions do estimates based on property descriptors \citep[e.g. conditional feature importance,][]{strobl2008conditional} differ from targeted learning approaches \citep[e.g. LOCO,][]{lei2018distribution,verdinelli2024decorrelated}? First works on this question \citep{verdinelli2024decorrelated,hooker2021unrestricted} indicate that targeted learning may be better suited for standard inferential tasks like estimating feature importance.

\paragraph{How to obtain realistic data?}
Many IML methods (e.g. Shapley values, LIME) rely on probing the ML model on modified instances \citep{scholbeck2019sampling}. These artificial ``data''  may be useful to audit the model, even though may never occur in the real world. However, if we want to learn about the world, artificial data is supposed to credibly supplement observations. Like others \citep{hooker2021bridging,hooker2021unrestricted,mentch2016quantifying}, we recommend respecting the dependency structure in the data if we strive to draw valid scientific inferences with IML.
\nl
However, obtaining realistic data is hard. Strategies such as our grade jitter strategy are useful, but require expert domain knowledge of the dependency structure in the data. Conditional density estimation techniques \citep[e.g. probabilistic circuits][]{choi2020probabilistic} or generative models (e.g. generative adversarial networks, normalizing flows, variational autoencoders, etc.) provide paths to generate realistic data without presupposing expert knowledge. Unfortunately, they are often computationally intensive. Also, current IML software implementations often only offer marginal versions of IML methods that are unsuited as property descriptors. We urge the IML research community to provide efficient implementations of conditional samplers and integrate them into IML packages.

\paragraph{Can we use property descriptors to encode background knowledge?}
We showed how to use property descriptors to extract knowledge from models. However, the converse direction of incorporating knowledge into models is also central for scientific progress \citep{dwivedi2021knowledge,nearing2021role,razavi2021deep}. There are already approaches that allow to enforce monotonicity constraints \citep{chipman2022mbart} or sparsity in the training process \citep{martius2016extrapolation}. But also property descriptors can be used to constrain the set of allowable models.\footnote{Fixing sufficiently many property descriptions even allows to completely determine the model in the case of the FANOVA decomposition \citep{apley2020visualizing,hooker2004discovering}.} For example, we could promote specific property descriptions during training by modifying the loss function to penalize models that deviate from them. Such strategies are indeed common in the fairness literature, where loss functions are designed to optimize for certain fairness metrics, which can be seen as property descriptors (see \citep{pessach2022review} for an overview).

\paragraph{What about non-tabular data? }
For some data types, such as images, audio, or video data, it is extremely difficult to formalize the estimand only in terms of low-level features such as pixels or audio frequencies. To follow our approach, we need a translation of high-level concepts (e.g. objects in images or words in audio) that scientists can use to formulate their questions in terms of the low-level features (e.g. pixels or audio frequencies) that the model works with. Such translations are notoriously difficult to find, proposals either rely on labeled data to learn such representations \citep{jia2013visual,zaeem2021cause,koh2020concept} or discuss constraints to learn them via unsupervised learning \citep{bengio2013representation,scholkopf2021toward}. We think that such a translation between low-level features and high-level concepts is one of the most pressing problems of IML research.

\section{Conclusion}
\label{sec:conclusion}
Traditional scientific models were designed to satisfy elementwise representationality, allowing scientists to learn about Nature by direct inspection of model elements. While ML models trained for prediction do not satisfy elementwise representationality, they do offer a unique ability to represent complex phenomena by digesting enormous amounts of noisy, multivariate and even multimodal observations. We have shown that it is still possible to learn about the phenomenon using them: all we need to do is to interrogate the model with suitable property descriptors. Our approach provides philosophers, IML researchers and scientists with a novel philosophical perspective on scientific representation with ML models and a valuable methodology for gaining insight into phenomena from such models using interpretation methods.

\backmatter

\section*{Acknowledgments}

We are very significantly indebted to Tom Sterkenburg, Seth Axen, Thomas Grote, Alexandra Gessner, Nacho Molina and Christian Scholbeck for their comments on the manuscript and their hints to related literature. We would also like to thank the two anonymous reviewers for their extremely valuable and constructive feedback, especially for pointing us to the indeed highly connected targeted learning framework.


\section*{Funding}
This work was supported by the Carl Zeiss stiftung (Project: Certification and Foundations of Safe Machine Learning Systems in Healthcare) and by the Deutsche Forschungsgemeinschaft (DFG, German Research Foundation) under Germany’s Excellence Strategy – EXC number 2064/1 – Project number 390727645.


%

\newpage
\bibliography{bibliography}
\newpage

\begin{appendices}

\section{Dataset}
\label{app:dataset}
\begin{figure}[h]
\centering
  \includegraphics[width=0.9\linewidth]{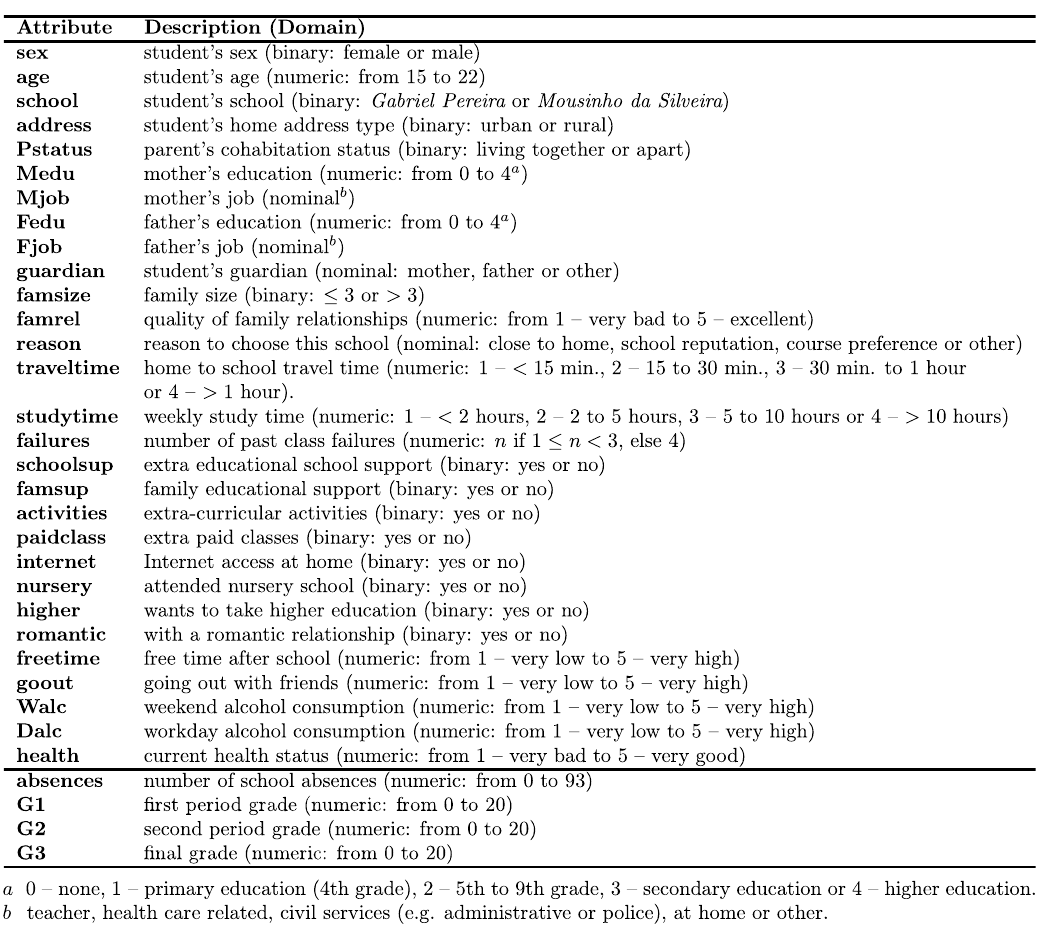}
  \caption{\textbf{Attributes in the \cite{cortez2008using} dataset}.}
  \label{fig:dataDesciption}
\end{figure}

\Cref{fig:dataDesciption} gives a descriptions of the different features and is copied from \cite{cortez2008using}. In our trained models, we only used the final G3 student grades. The data was collected during 2005 and 2006 from two public schools, from the Alentejo region in Portugal. The database is collected from a variety of sources from both school
reports and questionnaires. \cite{cortez2008using} integrated the information into a mathematics dataset (with 395 examples) and a Portuguese language dataset (649 records).

\end{appendices}

\end{document}